\documentclass[sigconf]{acmart}

\usepackage{booktabs} % For formal tables

\usepackage{multirow,tabularx}
\usepackage{hyperref}
\graphicspath{ {./images/} }
\usepackage{boldline}
\usepackage{algorithmic}
\usepackage{algorithm}
\usepackage{amsmath}
\usepackage{amssymb}

\usepackage{amsthm}

\theoremstyle{definition}
\newtheorem{definition}{Definition}[section]

\copyrightyear{2019}
\acmYear{2019} 
\setcopyright{iw3c2w3}
\acmConference[WWW '19]{Proceedings of the 2019 World Wide Web Conference}{May 13--17, 2019}{San Francisco, CA, USA}
\acmBooktitle{Proceedings of the 2019 World Wide Web Conference (WWW'19), May 13--17, 2019, San Francisco, CA, USA}
\acmPrice{}
\acmDOI{10.1145/3308558.3313504}
\acmISBN{978-1-4503-6674-8/19/05}
\usepackage{balance}
\newcommand{\customfootnotetext}[2]{{% Group to localize change to footnote
  \renewcommand{\thefootnote}{#1}% Update footnote counter representation
  \footnotetext[0]{#2}}}% Print footnote text

\begin{document}

\title{Stereotypical Bias Removal for Hate Speech Detection Task using Knowledge-based Generalizations}
%\titlenote{Produces the permission block, and
%  copyright information}
%\subtitle{Extended Abstract}
%\subtitlenote{The full version of the author's guide is available as
%  \texttt{acmart.pdf} document}

\author{Pinkesh Badjatiya, Manish Gupta*, Vasudeva Varma}
\affiliation{%
	\institution{IIIT Hyderabad}
	\city{Hyderabad}
	\country{India}
}
\email{pinkesh.badjatiya@research.iiit.ac.in, {manish.gupta,vv}@iiit.ac.in}


\begin{abstract}
With the ever-increasing cases of hate spread on social media platforms, it is critical to design abuse detection mechanisms to pro-actively avoid and control such incidents. While there exist methods for hate speech detection, they stereotype words and hence suffer from inherently biased training. Bias removal has been traditionally studied for structured datasets, but we aim at bias mitigation from unstructured text data. 

In this paper, we make two important contributions. First, we systematically design methods to quantify the bias for any model and propose algorithms for identifying the set of words which the model stereotypes. Second, we propose novel methods leveraging knowledge-based generalizations for bias-free learning. %Additionally, we evaluate the performance of our methods on multiple metrics to quantify the bias for any model and propose algorithms for identifying the set of terms which the model is biased on.

Knowledge-based generalization provides an effective way to encode knowledge because the abstraction they provide not only generalizes content but also facilitates retraction of information from the hate speech detection classifier, thereby reducing the imbalance. We experiment with multiple knowledge generalization policies and analyze their effect on general performance and in mitigating bias. Our experiments with two real-world datasets, a Wikipedia Talk Pages dataset (WikiDetox) of size $\sim$96k and a Twitter dataset of size $\sim$24k, show that the use of knowledge-based generalizations results in better performance by forcing the classifier to learn from generalized content. Our methods utilize existing knowledge-bases and can easily be extended to other tasks.

%We experiment report results for different bias detection as well as removal techniques. Our techniques show improved performance as compared to the state-of-the-art method.

%Additionally, we propose an automated technique to identify the sources of bias.
\end{abstract}


\newcommand{\tabitem}{~~\llap{\textbullet}~~}

\maketitle
\customfootnotetext{*}{Author is also a Principal Applied Researcher at Microsoft. (gmanish@microsoft.com)}

\section{Introduction}
%Why is this problem/domain critical

\begin{table}[!pbt]%
    \centering
    %\small
    \begin{tabular}{|p{0.5\linewidth}|p{0.21\linewidth}|p{0.11\linewidth}|}
        \hline
        \textbf{Examples} & \textbf{Predicted Hate Label (Score)}&\textbf{True Label}\\ \hline
        those guys are nerds & Hateful (0.83) & \multirow{10}{*}{Neutral}\\ \cline{1-2}
        can you throw that garbage please & Hateful (0.74) & \\ \cline{1-2}
        People will die if they kill Obamacare & Hateful (0.78) & \\ \cline{1-2}
        oh shit, i did that mistake again & Hateful (0.91) & \\ \cline{1-2}
        that arab killed the plants & Hateful (0.87)& \\  \cline{1-2}
        I support gay marriage. I believe they have a right to be as miserable as the rest of us. & Hateful (0.77) & \\
        
        \hline
        
        %        \hline
        %        christians and indians have to be swept clean so the americans can survive. & Neutral (0.35) & \multirow{4}{*}{Hateful}\\ \cline{1-2}
        %        Children should not go to school & Neutral (0.31) & \\ \cline{1-2}
        %        "I am colored, Latino and some Jewish. When I move in, I don't ruin the neighborhood, I wipe it out!" & Neutral (0.55) & \\ \cline{1-2}
        %        Chinese cannot write proper English& Neutral (0.39) & \\ \hline
    \end{tabular}
    \caption{Examples of Incorrect Predictions from the Perspective API}
    \label{tab:examples}
\end{table}

%Types of bias in the online world, %Impact of this bias
With the massive increase in user-generated content online, there has also been an increase in automated systems for a large variety of prediction tasks. While such systems report high accuracy on collected data used for testing them, many times such data suffers from multiple types of unintended biases. A popular unintended bias, stereotypical bias (SB), can be based on typical perspectives like skin tone, gender, race, demography, disability, Arab-Muslim background, etc. or can be complicated combinations of these as well as other confounding factors. Not being able to build unbiased prediction systems can lead to low-quality unfair results for victim communities. Since many of such prediction systems are critical in real-life decision making, this unfairness can propagate into government/organizational policy making and general societal perceptions about members of the victim communities~\cite{hovy2016social}.

%What is the actual problem
In this paper, we focus on the problem of detecting and removing such bias for the hate speech detection task. With the massive increase in social interactions online, there has also been an increase in hateful activities. Hateful content refers to ``posts'' that contain abusive speech targeting individuals (cyber-bullying, a politician, a celebrity, a product) or particular groups (a country, LGBT, a religion, gender, an organization, etc.). Detecting such hateful speech is important for discouraging such wrongful activities. While there have been systems for hate speech detection~\cite{badjatiya2017deep,waseem2016hateful,Djuric2015,kwok2013locate}, they do not explicitly handle unintended bias.

We address a very practical issue in short text classification: that the mere presence of a word may (sometimes incorrectly) determine the predicted label. We thus propose methods to identify if the model suffers from such an issue and thus provide techniques to reduce the undue influence of certain keywords.

%Poster child example
Acknowledging the importance of hate speech detection, Google launched Perspective API~\footnote{\url{https://www.perspectiveapi.com/}} in 2017. Table~\ref{tab:examples} shows examples of sentences that get incorrectly labeled by the API (as on $15^{th}$ Aug 2018). We believe that for many of these examples, a significant reason for the errors is bias in training data. Besides this API, we show examples of stereotypical bias from our datasets in Table~\ref{tab:examples2}.

\renewcommand{\tabcolsep}{2pt}
\begin{table*}[t!]
\centering
\small
\begin{tabular}{|p{5.2cm}|p{3.8cm}|p{9cm}|}
  \hline
  Sample & Classification prob. (Prediction) & Inference \\
%  \hline
%  bitch & 0.69 (sexist) & \multirow{6}{*}{ skdjfl skdjf lskjdf sjldf}\\
%   he & 0.84 (none) & \\
%   he is a bitch & 0.69 (sexist) & \\
%   she is a bitch & 0.69 (sexist) & \\
%   kat is a bitch & 0.92 (sexist) & \\
%   alice is a bitch & 0.72 (sexist) & \\
  \hline
%  kat & 0.43 (sexist) & \multirow{7}{*}{ skdjfl skdjf lskjdf sjldf}\\
%   women & 0.73 (none) & \\
%   islam & 0.73 (none) & \\
   kat is a woman & 0.63 (Hateful) & \multirow{2}{\linewidth}{Replacing the name from Kat to Alice should not have resulted in such a difference in prediction probability.}\\
      alice is a woman & 0.23 (Neutral) & \\
%   kat follows islam & 0.69 (none) & \\
   %kat works hard & 0.39 (sexist) & \\
%  \hline
 % kat & 0.43 (sexist) & \multirow{3}{*}{ skdjfl skdjf lskjdf sjldf}\\
  % feminist & 0.62 (sexist) & \\
   %kat is a feminist & 0.90 (sexist) & \\
  \hline
  that's just fucking great, man. & 0.80 (Hateful) & \multirow{3}{\linewidth}{Use of the word ``fucking'' somehow makes the sentence hateful. Similarly, use of the word ``shit'' abruptly increases the hate probability.} \\
    that's just sad man. & 0.24 (Neutral) & \\
    shit, that's just sad man. & 0.85 (Hateful) & \\
  \hline
  rt @ABC: @DEF @GHI @JKL i don't want to fit anything into their little boxes ... & 0.80 (Hateful) & In the training dataset, the word ``@ABC'' is mentioned in 17 tweets of which 13  (76\%) are ``Hateful''. Hence, the classifier associates @ABC with ``Hateful''\\
  %, which is a good enough solution, to begin with. To rectify this bias, we mask its presence in the training dataset.\\
  %\hline
  %that is f-4es boi & 0.7 (toxic) & Perspective API -- Examples created from the words obtained using the Max Probability BSW detection strategy.\\
    \hline
    my house is dirty    & 0.86 (Hateful) & \multirow{3}{\linewidth}{The words ``dirty'' and ``gotta'' are present heavily in instances labeled as hateful in the training dataset. This resulted in the classifier learning to wrongfully infer the sentence as being hateful.} \\
    i gotta go    & 0.71 (Hateful) & \\
    damn, my house is dirty. i gotta go and clean it & 0.99 (Hateful) & \\
    \hline
    \end{tabular}
    \caption{Examples of Incorrect Predictions from our Datasets described in Section~\ref{sec:datasets}}
    \label{tab:examples2}
\end{table*}

%The dissimilar presence of the sentences, as shown in Table \ref{tab:improper_frequency_of_samples}, in the observed dataset does not mean that both the sentences have different offensive tone. This again brings the point of a fair algorithm that treats all person and organizations equally irrespective of the caste, origin, and nature.

%Why related work is not enough
Traditionally bias in datasets has been studied from a perspective of detecting and handling imbalance, selection bias, capture bias, and negative set bias~\cite{khosla2012undoing}. In this paper, we discuss the removal of stereotypical bias. While there have been efforts on identifying a broad range of unintended biases in the research and practices around social data~\cite{Olteanu:2018:CRO:3159652.3162004}, there is hardly any rigorous work on principled detection and removal of such bias. One widely used technique for bias removal is sampling, but it is challenging to obtain instances with desired labels, especially positive labels, that help in bias removal. These techniques are also prone to sneak in further bias due to sample addition. To address such drawbacks, we design \emph{automated} techniques for bias detection, and \emph{knowledge-based generalization} techniques for bias removal with a focus on hate speech detection.

%+We design techniques for biased word detection.
%+Beyond that we propose techniques for detection as well as handling bias.
%+Bias removal for hate speech detection.

%-Data augmentation~cite{dixon2017measuring}
%how to get neutral sentences? Can induce bias for other terms.
%+Method for bias removal -- knowledge generalization

%What is your approach
    %General pipeline
    %Detection strategies
    %Replacement strategies
Enormous amounts of user-generated content is a boon for learning robust prediction models. However, since stereotypical bias is highly prevalent in user-generated data, prediction models implicitly suffer from such bias~\cite{caliskan2017semantics}. To de-bias the training data, we propose a two-stage method: Detection and Replacement. In the first stage (Detection), we propose skewed prediction probability and class distribution imbalance based novel heuristics for bias sensitive word (BSW) detection. Further, in the second stage (Replacement), we present novel bias removal strategies leveraging knowledge-based generalizations. These include replacing BSWs with generalizations like their Part-of-Speech (POS) tags, Named Entity tags, WordNet~\cite{Miller:1995:WLD:219717.219748} based linguistic equivalents, and word embedding based generalizations.

%Experimental details in short
We experiment with two real-world datasets -- WikiDetox and Twitter of sizes $\sim$96K and $\sim$24K respectively. The goal of the experiments is to explore the trade-off between hate speech detection accuracy and stereotypical bias mitigation for various combinations of detection and replacement strategies. Since the task is hardly studied, we evaluate using an existing metric Pinned AUC, and also propose a novel metric called PB (Pinned Bias). Our experiments show that the proposed methods result into proportional reduction in bias with an increase in the degree of generalization, without any significant loss in hate speech detection task performance.

%Contributions
In this paper, we make the following important contributions.
\begin{itemize}
\item We design a principled two-stage framework for stereotypical bias mitigation for hate speech detection task.
\item We propose novel bias sensitive word (BSW) detection strategies, as well as multiple BSW replacement strategies and show improved performance on existing metrics.
\item We propose a novel metric called PB (Pinned Bias) that is easier to compute and is effective.
\item We perform experiments on two large real-world datasets to show the efficacy of the proposed framework. % by tabulating experimental results across multiple metrics.
\end{itemize}

%Paper organization
The remainder of the paper is organized as follows. In Section~\ref{sec:related}, we discuss related work in the area of bias mitigation for both structured and unstructured data. In Section~\ref{sec:problem}, we discuss a few definitions and formalize the problem definition. We present the two-stage framework in Section~\ref{sec:approach}. We present experimental results and analysis in Section~\ref{sec:experiments}. Finally, we conclude with a brief summary in Section~\ref{sec:conclusions}.

\section{Related Work}
\label{sec:related}

Traditionally bias in datasets has been studied from a perspective of detecting and handling imbalance, selection bias, capture bias, and negative set bias~\cite{khosla2012undoing}. In this paper, we discuss the removal of stereotypical bias.

\subsection{Handling Bias for Structured Data}
These bias mitigation models require structured data with bias sensitive attributes (``protected attributes'') explicitly known. While there have been efforts on identifying a broad range of unintended biases in the research and practices around social data~\cite{Olteanu:2018:CRO:3159652.3162004}, only recently some research work has been done about bias detection and mitigation in the fairness, accountability, and transparency in machine learning (FAT-ML) community.  %Such work has mostly focused on (1) establishing the existence of bias~\cite{tatman2017gender,blodgett2017racial}, (2) handling bias in structured datasets with respect to ``protected attributes''~\cite{feldman2015certifying,beutel2017data,hardt2016equality}, or (3) handling bias for only particular stereotypes like gender~\cite{bolukbasi2016man}. 
While most of the papers in the community relate to establishing the existence of bias~\cite{tatman2017gender,blodgett2017racial}, some of these papers provide bias mitigation strategies by altering training data~\cite{feldman2015certifying}, trained models~\cite{hardt2016equality}, or embeddings~\cite{beutel2017data}. Kleinberg et al.~\cite{kleinberg2016inherent} and Friedler et al.~\cite{friedler2016possibility} both compare several different fairness metrics for structured data. Unlike this line of work, our work is focused on unstructured data. In particular, we deal with the bias for the hate speech detection task.

\subsection{Handling Bias for Unstructured Data}
Little prior work exists on fairness for text classification tasks. Blodgett and O'Connor~\cite{blodgett2017racial}, Hovy and Spruit~\cite{hovy2016social} and Tatman~\cite{tatman2017gender} discuss the impact of using unfair natural language processing models for real-world tasks, but do not provide mitigation strategies. Bolukbasi et al.~\cite{bolukbasi2016man} demonstrate gender bias in word embeddings and provide a technique to ``de-bias'' them, allowing these more fair embeddings to be used for any text-based task. But they tackle only gender-based stereotypes while we perform a very generic handling of various stereotypes. 

There is hardly any rigorous work on principled detection and removal of stereotypical bias from text data. Dixon et al.'s work~\cite{dixon2017measuring} is most closely related to our efforts. They handle stereotypical bias using a dictionary of hand-curated biased words, which is not generalizable and scalable. Also, one popular method for bias correction is data augmentation. But it is challenging to obtain instances which help in bias removal, and which do not sneak in further bias. To address such drawbacks, we design \emph{automated} techniques for bias detection, and \emph{knowledge-based generalization} techniques for bias removal with a focus on hate speech detection.

%+We design techniques for biased word detection.
%+Beyond that we propose techniques for detection as well as handling bias.
%+Bias removal for hate speech detection.

%-Data augmentation~cite{dixon2017measuring}
%how to get neutral sentences? Can induce bias for other terms.
%+Method for bias removal -- knowledge generalization

% \subsection{Hate Speech Detection}
% \todoPB{We can also talk about this paper: \href{https://qz.com/india/1333644/ibm-identifies-gender-bias-in-booker-prize-novel-shortlists/}{https://qz.com/india/1333644/ibm-identifies-gender-bias-in-booker-prize-novel-shortlists/}}

% The task of issues in production model is not new\cite{Bar-Ilan:2006:WLS:1165013.1165022}.
% The existence of bias in the learned algorithms is well known, especially in the image processing domain. Along with bias identification in image detection, recent works have attempted to debias the learning\cite{khosla2012undoing} in order to obtain models that perform well on cross-domain datasets.

% Recent works in the domain of NLP (Natural Language \& Processing) have identified similar bias in the existing machine learning/deep learning algorithms.

% \textcolor{red}{Links to all the PDFs: \href{https://www.dropbox.com/sh/bf64s7fxvekc15l/AAB2jv9LJvy7IZuoVchtwQL5a?dl=0}{URL}}

\section{Problem Formulation}
\label{sec:problem}
In this section, we present a few definitions and formulate a formal problem statement.

\subsection{Preliminary Definitions}
\label{sec:definitions}
We start by presenting definitions of a few terms relevant to this study in the context of hate speech detection task.

\begin{definition}{Hate Speech}: 
	Hate speech is speech that attacks a person or group on the basis of attributes such as race, religion, ethnic origin, national origin, sex, disability, sexual orientation, or gender identity. Its not the same as using certain profane words in text. A sentence can use profane words and still might not be hate speech. Eg, \textit{Oh shit! I forgot to call him}
\end{definition}

\begin{definition}{Hate Speech Detection}: 
Hate Speech Detection is a classification task where a classifier is first trained using labeled text data to classify test instances into one of the two classes: Hateful or Neutral.
\end{definition}

\begin{definition}{Stereotypical Bias (SB)}: 
In social psychology, a stereotype is an over-generalized belief about a particular category of people. However, in the context of hate speech detection, we define SB as an over-generalized belief about a word being Hateful or Neutral.
\end{definition}

For example, as shown in Table~\ref{tab:examples2}, ``kat'' and ``@ABC'' have a bias of being associated with ``Hateful'' class. Typical stereotype parameters represented as strings could also pertain to stereotypical bias in the context of hate speech detection, e.g. ``girls'', ``blacks'', etc.

\begin{definition}{Feature Units}: 
These are the smallest units of the feature space which are used by different text processing methods. 
\end{definition}

SB in the context of hate speech detection can be studied in terms of various feature units like character n-grams, word n-grams, phrases, or sentences. In this work, although we focus on words as feature units, the work can be easily generalized to other levels of text.

\begin{definition}{Bias Sensitive Word (BSW)}: 
A word $w$ is defined as a bias sensitive word for a classifier if the classifier is unreasonably biased with respect to $w$ to a very high degree. 
\end{definition}

For example, if a classifier labels the sentence with just a word ``muslims'' as ``Hateful'' with probability 0.81, then we say that the word ``muslims'' is a BSW for the classifier. Similarly, if a classifier predicts the sentence \textit{`Is Alex acting weird because she is a woman ?'} with much higher probability than \textit{`Is Alex acting weird because he is a man ?'} as Hateful, then the words ``woman'' and ``she'' are BSWs for that classifier.

\subsection{Problem Formulation}
\textbf{Given:} A hate speech detection classifier.\\
\textbf{Find:} 
\begin{itemize}
\item A list of BSW words from its training  data.
\item De-biased dataset to mitigate the impact of BSWs.
\end{itemize}

\begin{figure}%
\includegraphics[width=\columnwidth]{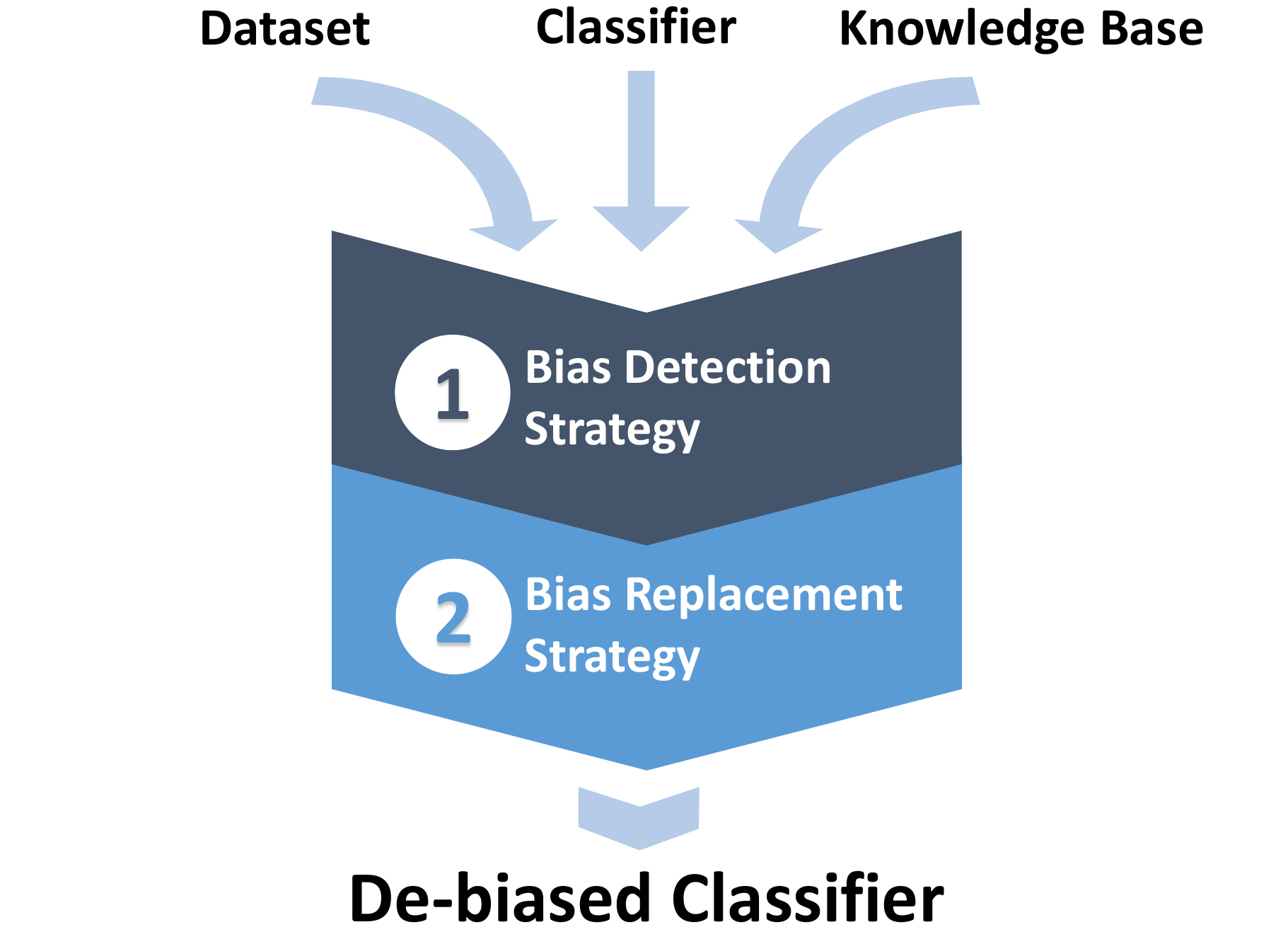}%
\caption{Conceptual Diagram for the proposed 2-Stage De-biasing Framework}%
\label{fig:arch}%
\end{figure}

We propose a two-stage framework to solve the problem. In the first stage (Detection), we propose biased-model prediction based novel heuristics for bias sensitive word (BSW) detection. Further, in the second stage (Replacement), we present novel bias removal strategies leveraging knowledge-based generalizations. Figure~\ref{fig:arch} shows the basic architecture diagram of the proposed framework.

% We take as input a classifier with an oracle access to its classification API and the FUs used by the classifier\\
% \textbf{Aim:} Once the bias is identified for a set of terms, we aim at reducing the bias by altering the training dataset. We do not change any stage of the training.\\
% \textbf{Solution:} For this reason we propose a way to,
% \begin{itemize}
%   \item Quantify the bias of the model for a given set of features
%   \item Algorithm to de-bias the dataset to assist bias-free learning
% \end{itemize}
%\input{debias_bias}
\section{Two-Stage De-biasing Approach}
\label{sec:approach}
In this section, we first discuss the first stage of the proposed framework which deals with detection of BSWs. Next, we discuss choices related to bias correction strategies that we could employ. Finally, as part of the second stage, we propose multiple knowledge generalization based BSW replacement strategies.

\subsection{Stage 1: Identifying BSWs}
\label{sec:bias_detection}

Many text-related tasks (e.g., gender profiling based on user-written text, named entity recognition) can extract strong signals from individual word occurrences within the text. However, higher order natural language processing tasks like hate speech detection, sentiment prediction, sarcasm detection cannot depend on strong signals from individual word occurrences; they need to extract signals from a combination of such words. Consider the sentence ``Pakistan is a country with beautiful mountains''. Here it is good for a named entity recognition classifier to use its knowledge about the significant association between the word ``Pakistan'' and a label ``location''. But it is not good for a hate speech detection classifier to use its knowledge about associating the word ``Pakistan'' with label ``Hateful'' which it might have unintendedly learned from the training data. 

Note that although it is important to identify BSWs, it is important to ensure that obvious hateful words should not be removed while doing bias mitigation. Using a standard abusive word dictionary, one can easily ignore such words from being considered a BSW. To identify stereotypical bias and avoid the classifier from propagating the bias from training data to its parameters, we propose the following three strategies for identifying BSWs. 

\subsubsection{Manual List Creation}
\citet{dixon2017measuring} manually created a set of 50 terms after performing analysis on the term usage across the training dataset. The selected set contains terms of similar nature but with disproportionate usage in hateful comments. Examples: lesbian, gay, bisexual, transgender, trans, queer, lgbt, etc. 

\subsubsection{Skewed Occurrence Across Classes (SOAC)} 
If a word is present a significant number of times in a particular class (say Hateful), the classifier is highly likely to learn a high probability of classifying a sentence containing that word into that particular class. Hence, we define the skewed occurrence measure to detect such BSWs.

For the hate speech detection task, consider the two classes: Hateful (+) vs Neutral (-). Let $tf(w)$ be the total term frequency of the word $w$ in the training dataset. Let $df(w)$ be the total document frequency, 
%$tf^{+}(w)$ be the term-frequency of the word for samples belonging to Hateful class, 
$df^{+}(w)$ be the document frequency of the word for documents labeled as Hateful, and $df^{-}(w)$ be the document frequency of the word for documents labeled as Neutral. 
%and similarly we define $tf^{-}(w)$ and $df^{-}(w)$.
Then, we can use the procedure mentioned in Algorithm~\ref{algo:BSW_skewed_occurrence} to rank the BSWs.

\begin{algorithm}[ht]
    \begin{algorithmic}[1]
        \STATE $S$ is a ordered list of words, $tf$ is  term-frequency of these words, $df$ is document-frequency values of these words, $\phi$ is the cutoff $tf$ threshold.
        \STATE $S^{*}\leftarrow\{w|w\in S~and~tf(w)>\phi~and~df^{+}(w) > df^{-}(w)\}$
        \STATE Sort $S^{*}$ in non-ascending order by ($df(w)$, $df^{+}(w)/ df(w)$).
%         $S^{*} \leftarrow [~]$ 
%         \FOR {$i$ in $1:length(S)$}
%             \IF {$df_{i}^{+} > df_{i}^{-}$ and $tf > \phi$}
%                 \STATE $S^{*} \leftarrow S^{*} + \{S_{i}\}$
%             \ENDIF
%         \ENDFOR

%         \FOR {$i$ in $1:length(S^{*})-1)$}
%         \FOR {$j$ in $(i+1):length(S^{*})$ }
%         \IF {$df_{i} < df_{j}$}
%             \STATE swap $S_{i}^{*}$ and $S_{j}^{*}$
%         \ELSE 
%             \STATE  $x \leftarrow df_{i}^{+} / df_{i}$
%             \STATE  $y \leftarrow df_{j}^{+} / df_{j}$
%             \IF {$x < y$}
%                 \STATE swap $S_{i}^{*}$ and $S_{j}^{*}$
%             \ENDIF
%         \ENDIF
%         \ENDFOR
%         \ENDFOR
        
    \end{algorithmic}
    \caption{BSW Detection based on `Skewed Occurrence Across Classes' detection strategy in Training Dataset}
    \label{algo:BSW_skewed_occurrence}
\end{algorithm}
Filtering based on just the high frequency would not be enough as that would include stopwords as well, hence we add additional constraints to filter these unnecessary terms.

Essentially, Algorithm~\ref{algo:BSW_skewed_occurrence} ranks a word higher if (1) it occurs with a minimum $\phi$ frequency in the entire dataset (2) it occurs in many training dataset documents (3) it occurs in Hateful documents much more than in Neutral documents. The drawback of this method is that we need to have access to the exact training corpus. 

\subsubsection{Skewed Predicted Class Probability Distribution (SPCPD)}
\label{detection_strat:skewed_probs}

Let $p(c|w)$ denote the classifier prediction probability of assigning a sentence containing only word $w$ to class $c$. Let $c_{\phi}$ denote a ``Neutral'' or an ``Others'' class, or any such ``catchall'' class. Then, we define the SPCPD score as the maximum probability of $w$ belonging to one of the classes excluding $c_{\phi}$.
    \begin{equation}
        SPCPD(w) = \max_{\forall c \in C, c \neq c_{\phi}} p(c|w)
    \end{equation}

Note that a high value of $SPCPD(w)$ means that the classifier has stereotyped the word $w$ to belong to the corresponding class. For example, if a classifier labels the sentence with just a word ``muslims'' as ``Hateful'' with probability 0.81, $SPCPD_{muslims}$=0.81. A word $w$ is called bias sensitive word (BSW) if $SPCPD(w)\geq \tau$, where $\tau$ is a threshold, usually close to 0.5 (for binary classification).

In settings where Hateful class is further sub-divided into multiple subtypes, using the maximum probability of classification amongst the positive target classes allows us to identify words that force the decision of the model towards a target class. Instead, an ideal classifier should predict a uniform probability distribution on the target classes except for the neutral class. The advantage of this bias detection strategy is that it does not need access to the original training dataset and can be used even if we just have access to the classifier prediction API. 

The $SOAC$ \& $SPCPD$ bias sensitive words detection strategies return an ordered list of BSWs with scores. We can pick the top few BSWs from each of these strategies, and subject them to bias removal as detailed in the remainder of this section.

Both the $SOAC$ \& $SPCPD$ bias sensitive words detection strategies can be used for n-grams as well. Considering that higher word-n-grams would require considerable computation power, we focus on unigrams only. Additionally, using prediction on single-word documents might seem trivial but later, in Section \ref{sec:experiments}, we show that such a simple intuitive strategy can be quite effective in identifying BSWs.

\subsection{Choices for Bias Correction Strategy}

For removing bias from a biased training dataset, we can do the following:
\begin{enumerate}
    \item
    \textbf{Statistical Correction:} This includes techniques that attempt to uniformly distribute the samples of every kind in all the target classes, augmenting the train set with samples to balance the term usage across the classes. ``Strategic Sampling'' is a simple statistical correction technique which involves strategically creating batches for training the classifier with aim of normalizing the training set. Data Augmentation is another technique that encourages adding new examples instead of performing sampling with replacement. 
    
    \textbf{Example:} Hatespeech corpus usually contains instances where `women' are getting abused in huge proportion as compared to `men' getting abused. Statistical correction can be useful in balancing this disproportion.
    
    \item
    \textbf{Data Correction:} This technique focuses on converting the samples to a simpler form by reducing the amount of information available to the classifier. This strategy is based on the idea of Knowledge generalization, when the classifier should be able to decide if the text is hateful or not based on the generalized sentence, not considering the private attribute information, as these attributes should not be a part of decision making of whether the sentence is hateful or not. Some popular examples of private attributes include Gender, Caste, Religion, Age, Name etc.
    
    \textbf{Example:} For a sentence, \textit{That Indian guy is an idiot and he owns a Ferrari}, private attribute information should not be used while learning to classify as abuse. The new generalized sentence after removing private attributes should look like: \textit{That <NATIONALITY> guy is an idiot and he owns a <CAR>}.
    
    \item
    \textbf{Model Correction or Post-processing:} Rather than correcting the input dataset or augmenting, one can either make changes to the model like modifying word embeddings or post-process the output appropriately to compensate for the bias.
    
    \textbf{Example:} A simple technique would involve denying predictions when the sentence contains private attributes to reduce false positives.
\end{enumerate}

In this paper, we direct our effort towards effective data correction based methods to counter bias in learning. This is mainly because of the following reasons: (1) for statistical correction, it is challenging to obtain instances which help in bias removal, and which do not sneak in further bias. In other words, the additional words in augmented instances can skew the distribution of existing words, inducing new bias. (2) for model correction, the strategy could be very specific to the type of classifier model.

We consider dataset augmentation or statistical correction as a baseline strategy. In order to prevent differential learning, we can add additional examples of neutral text in the training set containing the BSWs forcing the classifier to learn features which are less biased towards these BSWs. \citet{dixon2017measuring} use this strategy to balance the term usage in the train dataset thereby preventing biased learning.

\subsection{Stage 2: Replacement of BSWs}
\label{sec:bias_correction}

Next, we propose multiple knowledge generalization based BSW replacement strategies as follows.

\subsubsection{Replacing with Part-of-speech (POS) tags} 

In order to reduce the bias of the BSWs, we can replace each of these words with their corresponding POS tags. This process masks some of the information available to the model during training inhibiting the classifier from learning bias towards the BSWs.
    
\textbf{Example:} Replace the word `Muhammad' with POS tag `NOUN' in the sentence - \textit{Muhammad set the example for his followers, and his example shows him to be a cold-blooded murderer.} POS replacement substitutes specific information about the word `Muhammad' from the text and only exposes partial information about the word, denoted by its POS tag `NOUN', forcing it to use signals that do not give significant importance to the word `Muhammad' to improve its predictions.
    
\subsubsection{Replacing with Named-entity (NE) tags} 

Consider the sentence - \textit{Mohan is a rock star of Hollywood}. The probability of the sentence being hateful or not does not depend on the named entities `Mohan' and `Hollywood'. However, classifiers are prone to learn parameters based on the presence of these words. Hence, in this strategy, we replace the named entities with their corresponding entity type tag. This technique generalizes all the named-entities thus preventing information about the entities from being exposed to the classifier.
    
We use the standard set of NE tags which includes PERSON, DATE, PRODUCT, ORGANIZATION etc.

\subsubsection{K-Nearest Neighbour}

Given a word embedding space (say GloVe~\cite{pennington2014glove}) and a BSW $w$, we find $K$ nearest neighbors for $w$ in the embedding space. Let the $N_K(w)$ indicate the set of these $K$ nearest neighbors plus the word $w$. For every occurrence of $w$ in the training dataset, we randomly replace this occurrence with a word selected from $N_K(w)$. The intuition behind the strategy is as follows. If we replace all instances of a word with another word then it may not decrease the bias. But if we replace some sentences with other variants of the word, then it should add some variance in the sentences, and hence reduce the bias.

This strategy should serve as an equivalent of adding additional examples in the text to reduce the class imbalance.
    
\subsubsection{Knowledge generalization using lexical databases}
    
Replacing surface mentions of entities with their named entity types holds back important information about the content of the sentence; it generalizes the content too much. A more general approach would involve replacing BSWs with the corresponding generalized counterparts. A major reason for learning bias is because of differential use of similar terms. Consider the terms, `women' and `man'. For the domain of hate speech, the term `women' would be present more frequently as compared to `man', even though they represent similar, if not same, groups.
    
We leverage WordNet~\cite{Miller:1995:WLD:219717.219748} database to obtain semantically related words and the lexical relations in order to find a suitable candidate for replacement. In this strategy, words are replaced with their corresponding hypernym as obtained from the WordNet. A hypernym is a word with a broad meaning constituting a category into which words with more specific meanings fall; a superordinate. For example, \textit{colour is a hypernym of red.} A BSW might have multiple hypernyms and therefore different paths leading to a common root in WordNet. We perform a Breadth-First-Search (BFS) over the sets of synonyms, known as synsets, from the child node as obtained from the WordNet for a BSW. A synset contains lemmas, which are the base form of a word. We then iterate over all the lemmas in the synset to obtain the replacement candidate. This allows us to identify the lowest parent in the hypernym-tree which is already present in the training dataset vocabulary.
    
We try out all the words in the synonym-set to select a suitable candidate for replacement. Replacing a word with a parent which is not present in the text won't be of any importance since it will not help reduce bias when learning the classifier. Hence, we ignore such words. Not every word can be located in WordNet, especially the ones obtained from casual conversational forums like Twitter. To obtain better performance, we use spell correction before searching for the suitable synset in the WordNet.

\subsubsection{Centroid Embedding} 

All the above techniques exploit the information in the text space either by using NLP tools or lexical databases. In order to leverage the complex relationships in the language in embedding space, we can use the existing word embeddings which sufficiently accurately capture the relationship between different words.

In the centroid embedding method, we replace a bias sensitive word with a dummy tag whose embedding we compute as follows. We find the POS tag for that occurrence of the word, and then find similar words from the word embeddings with similar POS usage. Further, we compute the centroid of the top $k$ neighbors, including the original word to obtain the centroid embedding. We set $k$ to 5 in our experiments.
    
The centroid represents the region of the embedding space to which the bias sensitive word belongs. So, rather than using the exact point embedding of the word, we smooth it out by using the average embedding of the neighborhood in which the point lies.
%Replacing with its centroid embedding tag results in words closer in the embedding space to have similar embedding representation, if they had similar POS usage. 
%This replacement strategy encourages word with similar semantic and POS usage to be further closer in the embedding space. 
This strategy is a continuous version of the WordNet-based strategy as the generalizations happen in the embedding space rather in text or word space. Additionally, computing the centroid of sufficient number of words helps in obtaining bias-free vector representations giving improved performance (as shown in Section~\ref{sec:experiments}).

\section{Experiments}
\label{sec:experiments}

\bgroup
\begin{table*}[!htb]
    \centering
    \small
    \begin{tabular}{|p{3cm}|p{7cm}|p{7cm}|}
        \hline
        \multicolumn{1}{|c|}{\textbf{Detection Strategy}} &\multicolumn{1}{c|}{\bf{WikiDetox Dataset}} & \multicolumn{1}{c|}{\bf{Twitter Dataset}} \\\hline
        %    & \multicolumn{1}{c|}{\bf{Wikipedia Dataset}} & \multicolumn{1}{c|}{\bf{Twitter Dataset}} \\\hline/
        Skewed Occurrence Across Classes & get, youre, nerd, mom, yourself, kiss, basement, cant, donkey, urself, boi & lol, yo, my, ya, girl, wanna, gotta, dont, yall, yeah \\\hline
        Skewed Predicted Class Probability Distribution & pissed, gotta, sexist, righteous, kitten, kidding, snake, wash, jew, dude & feelin, clap, lovin, gimme, goodnight, callin, tryin, screamin, cryin, doin \\\hline \hline
        
        \multicolumn{1}{|c|}{\textbf{Detection Strategy}}    & \multicolumn{2}{c|}{\textbf{WikiDetox Dataset}} \\ \hline
        Manually curated set of words & \multicolumn{2}{p{14cm}|}{lesbian, gay, bisexual, transgender, trans, queer, lgbt, lgbtq, homosexual, straight, heterosexual, male, female, nonbinary, african, african american, black, white, european, hispanic, latino, latina, latinx, mexican, canadian, american, asian, indian, middle eastern, chinese, japanese, christian, muslim, jewish, buddhist, catholic, protestant, sikh, taoist, old, older, young, younger, teenage, millenial, middle aged, elderly, blind, deaf, paralyzed} \\\hline
    \end{tabular}
    
    %   \begin{tabular}{|p{3cm}|p{10.1cm}|}
    %     \hline
    %      & \textbf{Wikipedia Dataset} \\ \hline
    %     Manually curated set of words & lesbian, gay, bisexual, transgender, trans, queer, lgbt, lgbtq, homosexual, straight, heterosexual, male, female, nonbinary, african, african american, black, white, european, hispanic, latino, latina, latinx, mexican, canadian, american, asian, indian, middle eastern, chinese, japanese, christian, muslim, jewish, buddhist, catholic, protestant, sikh, taoist, old, older, young, younger, teenage, millenial, middle aged, elderly, blind, deaf, paralyzed \\\hline
    %   \end{tabular}
    
    \caption{BSWs obtained using the Bias Detection strategies on the WikiDetox and Twitter dataset. The words in the order from left to right are ranked based on the strategy specific ranking criteria.}
    \label{tab:biasedWords}
\end{table*}
\egroup

In this section, we first discuss our datasets and metrics. Next, we present comparative results with respect to hate speech detection accuracy and bias mitigation using different BSW detection and replacement strategies.

\subsection{Datasets}
\label{sec:datasets}
In order to measure the efficacy of the proposed techniques, we use the following datasets to evaluate the performance.

\textbf{WikiDetox Dataset~\cite{wulczyn2017ex}:} We use the dataset from the Wikipedia Talk Pages containing 95,692 samples for training, 32,128 for development and 31,866 for testing. Each sample in the dataset is an edit on the Wikipedia Talk Page which is labeled as `Neutral' or `Hateful'.

\textbf{Twitter  Dataset~\cite{davidson2017automated}:} We use the dataset consisting of hateful tweets obtained from Twitter. It consists of 24,783 tweets labeled as hateful, offensive or neutral. We combine the hateful and offensive categories; and randomly split the obtained dataset into train, development, and test in the ratio 8:1:1 to obtain 19,881 train samples, 2,452 development samples, and 2,450 test samples.

\textbf{``Wiki Madlibs'' Eval Dataset~\cite{dixon2017measuring}:} For the Dataset augmentation based bias mitigation baseline strategy, we follow the approach proposed by~\citet{dixon2017measuring}. For every BSW, they manually create templates for both hateful instances and neutral instances. Further, the templates are instantiated using multiple dictionaries resulting into $\sim$77k examples with 50\% examples labeled as Hateful. For example, for a BSW word ``lesbian'', a template for a neutral sentence could be ``Being lesbian is $\langle$adjectivePositive$\rangle$.'' where $\langle$adjectivePositive$\rangle$ could be a word from dictionary containing ``great, wonderful, etc.'' A template for a Hateful sentence could be ``Being lesbian is $\langle$adjectiveNegative$\rangle$.'' where $\langle$adjectiveNegative$\rangle$ could be a word from dictionary containing ``disgusting, terrible, etc.''.

\subsection{Metrics}
\subsubsection{ROC-AUC Accuracy}
For the hate speech detection task, while we mitigate the bias, it is still important to maintain a good task accuracy. Hence, we use the standard ROC-AUC metric to track the change in the task accuracy.

\subsubsection{Pinned AUC Equality Difference}

In order to empirically measure the performance of the de-biasing procedures,~\citet{dixon2017measuring} proposed the Pinned AUC Equality Difference ($pAUC$) metric to quantify the bias in learning for the dataset augmentation strategy only. 

For a BSW $w$, let $D_{w}$ indicate the subgroup of instances in the ``Wiki Madlibs'' eval dataset containing the term $w$. Let $\overline{D_{w}}$ be the instances which do not contain the term $w$ such that $|D_{w}|=|\overline{D_{w}}|$. Further, the classifier is used to obtain predictions for the entire subgroup dataset $\{D_{w}\cup \overline{D_{w}}\}$. Let $AUC(w)$ denote the subgroup AUC obtained using these predictions. Further, let $AUC$ denote the AUC obtained using predictions considering the entire ``Wiki Madlibs'' dataset $\{\sum_{\forall w} (D_{w}\cup \overline{D_{w}})\}$. Then, a good way to capture bias is to look at divergence between subgroup AUCs and the overall AUC. \citet{dixon2017measuring} thus define the Pinned AUC Equality Difference ($pAUC$) metric as follows,

\begin{equation}
    pAUC = \sum_{w \in BSW} |AUC - AUC(w)|
    \label{eq:pinnedAUC}
\end{equation}

A lower $pAUC$ sum represents less variance between performance on individual term subgroups and therefore less bias. However, the $pAUC$ metric suffers from the following drawbacks:

% The pinned AUC metric for a subgroup is defined by computing the AUC on a secondary dataset containing two equally balanced components: a sample of comments from the subgroup of interest and a sample of comments that reflect the underlying distribution of comments. By creating this auxiliary dataset that ``pin's'' the subgroup to the underlying distribution, the AUC captures the divergence of the model performance on one subgroup with respect to the average example.
        
%     \begin{equation}
%         pD_{t} = s(D_{t}) + s(D), |s(D_{t})| = |s(D)|
%         \label{eq:pinnedAUC1}
%     \end{equation}
    
%     \begin{equation}
%         pAUC_{t} = AUC(pD_{t})
%         \label{eq:pinnedAUC2}
%     \end{equation}
    
%     Formally, let $D$ represent the full set of samples
%     and $D_{t}$ be the set of examples in subgroup $t$, then we
%     can generate the secondary dataset for term $t$ by applying
%     some sampling function $s$ as mentioned in Equation \ref{eq:pinnedAUC1}. Equation \ref{eq:pinnedAUC2} then defines the pinned AUC of term $t$, $pAUC_{t}$. 

%     Lower score is better. Using these definition we can compute the pinned AUC Equality  Difference as:
    
%     \begin{multline*}
%         Pinned~AUC~Equality~Difference = \sum_{t \in T} |AUC - pAUC_{t}|
%         \label{eq:pinnedAUC3}
%     \end{multline*}
        
    \begin{itemize}
        \item Computation of $pAUC$ requires creating a synthetic dataset with balanced examples for each bias sensitive word (BSW) $w$, making it challenging to compute and scale with the addition of new BSWs as it would require an addition of new examples with desired labels.
        \item The ``Wiki Madlibs'' Eval dataset is synthetically generated based on templates defined for BSWs. The pinning to the AUC obtained on such a dataset is quite ad hoc. An ideal metric should be independent of such relative pinning.
        % data and should reflect the error or distance from an ideal setting, not a hypothetical one encouraging the need for absolute evaluation measures.
        %\item pinnedAUC inherently captures the variation of the subgroup performance wrt the general performance not ideal case. For sensitive tasks such as hate speech, bias in the evaluation dataset can lead to a shifted probability distribution which the metric would not be able to capture. Even for cases when the subgroups are not of a similar nature, the metric would not be usable.
    \end{itemize}
    
\bgroup
\begin{table*}[!pbth]%
    \centering
    \small
    \begin{tabular}{|p{3cm}|r|r|r|r||r|r|r|r|}
        \hline
        \textbf{Detection \newline Strategy $\rightarrow$}  & \multicolumn{4}{c||}{Skewed Occurrence Across Classes} &   \multicolumn{4}{c|}{Skewed Predicted Class Prob. Distribution}\\
        \hline
        %\multirow{2}{\linewidth}{\textbf{Replacement Strategy $\downarrow$}} & ROC-AUC & $PB_{mean}$ & $PB_{sym}$ & $PB_{asym}$ & ROC-AUC & $PB_{mean}$ & $PB_{sym}$ & $PB_{asym}$\\
        \textbf{Replacement Strategy $\downarrow$} & ROC-AUC & $PB_{mean}$ & $PB_{sym}$ & $PB_{asym}$ & ROC-AUC & $PB_{mean}$ & $PB_{sym}$ & $PB_{asym}$\\
%        & & \multicolumn{1}{c|}{$(10^{-1})$} & \multicolumn{1}{c|}{$(10^{-1})$} & \multicolumn{1}{c||}{$(10^{-1})$} & & \multicolumn{1}{c|}{$(10^{-1})$} & \multicolumn{1}{c|}{$(10^{-1})$} & \multicolumn{1}{c|}{$(10^{-1})$} \\
        \hline
        Biased & 0.955 & 0.0174 & 0.4370 & 0.4370 & 0.958 & \underline{0.0848} & 0.2553 & 0.2553 \\\hline
        NER tags & \underline{0.961} & \underline{0.0140} & 0.4542 & 0.4542 & 0.957 & 0.1266 & 0.1869 & 0.1753 \\\hline
        POS tags & 0.960 & 0.1271 & \underline{0.1209} & \underline{0.0901} & \textbf{0.961} & 0.1235 & \underline{0.1478} & \underline{0.1110} \\\hline
        K-Nearest Neighbour & 0.957 & 0.1323 & 0.3281 & 0.3248 & \textbf{0.961} & 0.1311 & 0.2421 & 0.2370 \\\hline
        WordNet - 0 & 0.960 & 0.2238 & 0.3081 & 0.2906 & 0.954 & 0.1105 & 0.2079 & 0.2016 \\\hline
        WordNet - 1 & 0.961 & 0.2206 & 0.3167 & 0.2951 & 0.957 & 0.1234 & 0.2123 & 0.2016 \\\hline
        WordNet - 2 & 0.953 & 0.1944 & 0.3016 & 0.2870 & \textbf{0.961} & 0.1590 & 0.2059 & 0.1782 \\\hline
        WordNet - 3 & \textbf{0.963} & 0.1824 & 0.3420 & 0.3259 & \textbf{0.961} & 0.1255 & 0.1889 & 0.1787 \\\hline
        WordNet - 4 & 0.957 & 0.0960 & 0.3851 & 0.3831 & 0.956 & 0.1308 & 0.2053 & 0.1928 \\\hline
        WordNet - 5 & 0.956 & 0.0808 & 0.3909 & 0.3873 & 0.959 & 0.1089 & 0.1909 & 0.1844 \\\hline
        Centroid Embedding & 0.958 & \textbf{0.0000} & \textbf{0.0204} & \textbf{0.0000} & \underline{0.960} & \textbf{0.0709} & \textbf{0.0578} & \textbf{0.0439} \\\hline
    \end{tabular}
    \caption{Hate Speech Detection Accuracy and Bias Mitigation Results Across Multiple Detection and Replacement Strategies on Twitter Dataset. The best replacement strategy is highlighted in \textit{bold} while the $2^{nd}$ best is \textit{\underline{underlined}} for each metric.}
    \label{tab:TwitterResult2}
\end{table*}
\egroup

\subsubsection{Pinned Bias (PB)}

To address the above-mentioned drawbacks with $pAUC$, we propose a family of metrics 
%, similar to mean absolute error (MAE), 
called Pinned Bias Metric (PB) that is better able to capture the stereotypical bias. Note that small values of PB are preferred.

Let $T$ represent the set of BSW words. Let $p(``Hateful\text{''}|w)$ be the prediction probability for a sentence which contains only word $w$. We define $PB$ as the absolute difference in the predictions `pinned' to some predictive value for the set $T$. This captures the variation in Bias Probability of the classifier for the given set of words with respect to the pinned value. Keeping these intuitions in mind, we define the general PB metric as follows. 

\begin{equation}
    PB=\sum_{w\in T} 
        \frac{|p(``Hateful\text{''}|w)-\phi|}{|T|}
    \label{eq:PB}
\end{equation}

\noindent, where $\phi$ is the pinned value which differs for different metrics in the PB family. 
%and $N$ is the normalizing factor. 
Next, we define three specific members of the PB family as follows.

\begin{itemize}
\item $PB_{mean}$: For a similar set of terms, it is important to have consistency in the predictions. To capture this deviation in prediction from the general classifier consensus, we pin the metric with respect to the mean prediction. For this metric, $\phi$ in Eq.~\ref{eq:PB} is set as follows.

\begin{equation}
\phi=\sum_{w\in T} \frac{p(``\operatorname{Hateful}\text{''}|w)}{|T|}
\end{equation}

\item $PB_{sym}$: $PB_{mean}$ is not effective in situations where the terms are of varied nature or from diverse domains. Also pinning it relative to the general consensus of the classifier again can introduce errors.

Hence, for this metric, $\phi$ in Eq.~\ref{eq:PB} is set as $\phi$=0.5 for binary classification. In case of $k$ classes, $\phi$ can be set to $\frac{1}{k}$.

\item $PB_{asym}$: 
%The prediction probabilities are not symmetric on the prediction scale of 0-1 encouraging us to inherit this asymmetrical nature in the metric. 
Intuitively, prediction probability of 0.70 being hateful as compared to probability of 0.30 being hateful for a BSW should not be considered equal as a degree of being hateful is much more sensitive as compared to a degree of being neutral.

A high value of $p(``Hateful\text{''}|w)$ should be penalized much more than a value less than 0.50 (for binary classification; $\frac{1}{k}$ for $k$-class classification). Hence, for this metric, $\phi$ in Eq.~\ref{eq:PB} is set as follows.
\begin{equation}
\phi=\min(p(``Hateful\text{''}|w),~0.5)
\end{equation}
\end{itemize}

The $PB$ metrics, in general, represent the average deviation of the probability scores from the ideal pinning value. Eg, A score of $PB_{sym} = 0.2$ would mean that the probabilities have an average deviation of 0.2 from the ideal value of 0.5.  The $PB$ metrics can be modified, if needed, to incorporate the bias in n-grams.

\subsection{Experimental Settings}
\begin{figure*}%
	\includegraphics[width=\linewidth]{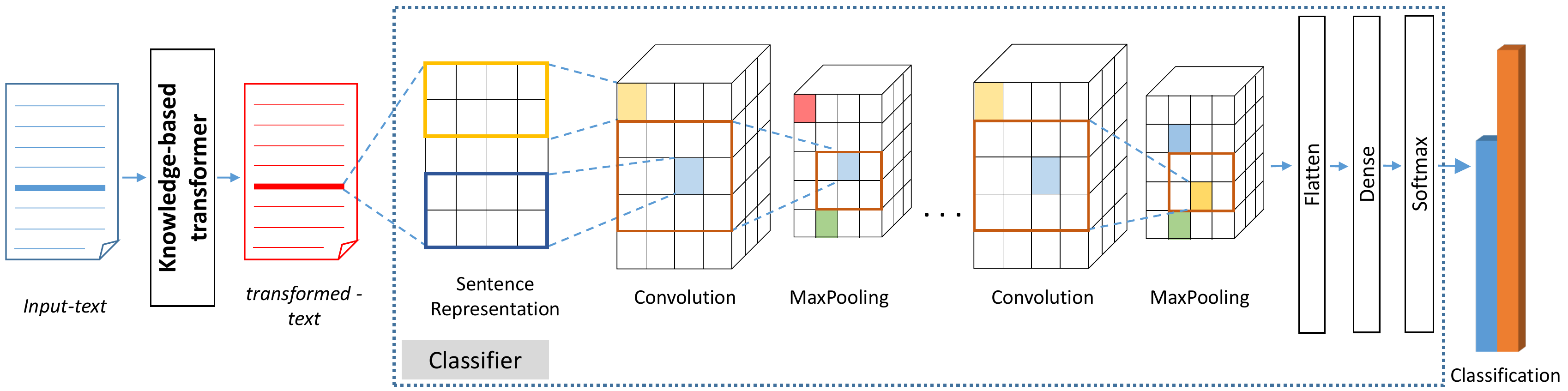}%
	\caption{Model Architecture Diagram}%
	\label{fig:modelArchitecture}%
\end{figure*}

We perform multiple experiments to test different BSW identification and bias mitigation strategies. We experiment with multiple classification methods like Logistic Regression, Multi-layer Perceptron (MLP), Convolutional Neural Network (CNN), an ensemble of classifiers. Since we obtained similar results irrespective of the classification method, we report results obtained using CNNs in this section.

We use identical Convolutional Neural Network (CNN)~\cite{dixon2017measuring} implemented in Keras~\cite{chollet2015keras} for all the experiments. Fig.~\ref{fig:modelArchitecture} shows the architecture diagram of the neural network. We use 100D GloVe~\cite{pennington2014glove} embeddings to encode text and CNN architecture with 128 filters of size $5\times 5\times 5$ for each convolution layer, dropout of 0.3, max-sequence length as 250, categorical-cross-entropy as the loss function and RMSPROP as optimizer with learning-rate of 0.00005. We use the best parameter setting as described in \citet{dixon2017measuring} and train the classifiers for our task with batch size 128 and early stopping to a maximum of 20 epochs, keeping other hyperparameters constant throughout the experiments.

We use the spaCy~\footnote{https://spacy.io/} implementation of POS tagger and NER tool. More details about the experiments can be found in the source code~\footnote{http://bit.ly/stereotypical-bias-removal}.
%We train 5 identical CNN models for each strategy and report average scores to prevent variance due to training. For comparison with previous results, we use 10 variations of each model in Manual Words set + Wikipedia dataset with results tabulated in Table \ref{tab:different_strategies_results}.

%A major cause of bias towards words is due to the fact how CNNs work. They rely on Max pooling, hence giving differential importance to a set of features. We are able to identify these by using the words as input to the model in order to identify the set of features that the model is biased on. Ideally, it should be biased on a combination of features, but that would increase the number of combinations that we need to explore exponentially. We only experiment with individual words.

\subsection{Results}

\bgroup
\begin{table}[!phtb]
    \centering
    \small
    \begin{tabular}{|l|r|r|}
        \hline
        & ROC-AUC & $pAUC$ \\\hline
        Biased & 0.948 & 2.785\\\hline
        Wiki De-bias~\cite{dixon2017measuring} & 0.957 & 1.699\\\hline
        NER tags & 0.945 & 2.893\\\hline
        POS tags & \underline{0.958} & \underline{1.053}\\\hline
        K-Nearest Neighbour & 0.910 & 3.480\\\hline
        WordNet - 0 & 0.943 & 2.781\\\hline
        WordNet - 1 & 0.949 & 2.530\\\hline
        WordNet - 2 & \underline{0.958} & 1.634\\\hline
        WordNet - 3 & 0.954 & 2.068\\\hline
        WordNet - 4 & 0.951 & 2.390\\\hline
        WordNet - 5 & 0.954 & 1.992\\\hline
        Centroid Embedding & \textbf{0.971} & \textbf{0.629}\\\hline
    \end{tabular}
    \caption{Bias Mitigation Results using pAUC metric and ROC-AUC across Multiple Replacement Strategies on Wiki Madlibs Dataset. The best replacement strategy is highlighted in \textit{bold} while the $2^{nd}$ best is \textit{\underline{underlined}} for each metric.}
    \label{tab:madlibsResult3}
\end{table}
\egroup

\bgroup
\begin{table*}[!htb]%
    \centering
    \small
    \begin{tabular}{|p{2.5cm}|r|r|r|r||r|r|r|r||r|r|r|r|}
        \hline
        \textbf{Detection \newline Strategy $\rightarrow$} & \multicolumn{4}{c||}{Manual List Creation} & \multicolumn{4}{c||}{Skewed Occurrence Across Classes} &   \multicolumn{4}{c|}{Skewed Predicted Class Prob. Distribution}\\
        \hline
        %\multirow{2}{\linewidth}{\textbf{Replacement Strategy $\downarrow$}} & ROC-AUC & $PB_{mean}$ & $PB_{sym}$ & $PB_{asym}$ & ROC-AUC & $PB_{mean}$ & $PB_{sym}$ & $PB_{asym}$ & ROC-AUC & $PB_{mean}$ & $PB_{sym}$ & $PB_{asym}$\\
        %& & \multicolumn{1}{c|}{$(10^{-1})$} & \multicolumn{1}{c|}{$(10^{-1})$} & \multicolumn{1}{c||}{$(10^{-1})$} & & \multicolumn{1}{c|}{$(10^{-1})$} & \multicolumn{1}{c|}{$(10^{-1})$} & \multicolumn{1}{c||}{$(10^{-1})$} & & \multicolumn{1}{c|}{$(10^{-1})$} & \multicolumn{1}{c|}{$(10^{-1})$} & \multicolumn{1}{c|}{$(10^{-1})$} \\
        \multicolumn{1}{|c|}{\textbf{Replacement Strategy} $\downarrow$} & ROC-AUC & $PB_{mean}$ & $PB_{sym}$ & $PB_{asym}$ & ROC-AUC & $PB_{mean}$ & $PB_{sym}$ & $PB_{asym}$ & ROC-AUC & $PB_{mean}$ & $PB_{sym}$ & $PB_{asym}$\\
        \hline
        Biased & \underline{0.957} & 0.0806 & 0.1223 & 0.0144 & 0.952 & 0.1424 & \underline{0.1473} & 0.0666 & 0.956 & 0.1826 & 0.2101 & 0.0505\\\hline
        Wiki Debias~\cite{dixon2017measuring} & \textbf{0.958} & 0.0491 & 0.1433 & 0.0058 & NA & NA & NA & NA & NA & NA & NA & NA\\\hline
        NE tags & 0.956 & 0.0833 & 0.1269 & 0.0137 & \textbf{0.959} & 0.1741 & 0.2252 & 0.0421 & 0.955 & 0.1533 & 0.2118 & 0.0304\\\hline
        POS tags & 0.953 & \underline{0.0251} & \underline{0.0540} & \underline{0.0020} & \underline{0.954} & 0.1614 & 0.2033 & 0.0481 & \textbf{0.958} & 0.0789 & 0.3656 & \textbf{0.0000}\\\hline
        K-Nearest Neighbour & 0.937 & 0.0554 & 0.1120 & 0.0023 & 0.936 & \textbf{0.1154} & \textbf{0.1196} & 0.0431 & 0.934 & \textbf{0.0618} & \textbf{0.1284} & 0.0035\\\hline
        Wordnet - 0 & 0.956 & 0.0765 & 0.1188 & 0.0121 & 0.946 & 0.1445 & 0.1620 & 0.0435 & 0.951 & 0.1835 & 0.2602 & 0.0283\\\hline
        Wordnet - 1 & 0.956 & 0.0756 & 0.1201 & 0.0133 & \underline{0.954} & 0.1820 & 0.2132 & 0.0529 & \textbf{0.958} & 0.1191 & 0.3401 & \underline{0.0002}\\\hline
        Wordnet - 2 & 0.956 & 0.0735 & 0.1182 & 0.0110 & 0.952 & 0.1767 & 0.1986 & 0.0550 & 0.955 & 0.1453 & 0.3281 & 0.0037\\\hline
        Wordnet - 3 & 0.953 & 0.0708 & 0.1151 & 0.0098 & 0.941 & \underline{0.1297} & 0.1622 & \textbf{0.0326} & 0.952 & 0.1699 & 0.3029 & 0.0121\\\hline
        Wordnet - 4 & 0.953 & 0.0733 & 0.1133 & 0.0097 & 0.946 & 0.1398 & 0.1570 & \underline{0.0393} & 0.955 & 0.1751 & 0.3150 & 0.0123\\\hline
        Wordnet - 5 & 0.948 & 0.0653 & 0.1084 & 0.0057 & \textbf{0.959} & 0.1766 & 0.2336 & 0.0468 & 0.955 & 0.1561 & 0.3475 & 0.0060\\\hline
        Centroid Embedding & 0.955 & \textbf{0.0000} & \textbf{0.0499} & \textbf{0.0000} & 0.953 & 0.1803 & 0.1918 & 0.0535 & \underline{0.957} & \underline{0.0774} & \underline{0.1949} & \textbf{0.0000}\\\hline
        
    \end{tabular}
    \caption{Hate Speech Detection Accuracy and Bias Mitigation Results Across Multiple Detection and Replacement Strategies on WikiDetox Dataset. The best replacement strategy is highlighted in \textit{bold} while the $2^{nd}$ best is \textit{\underline{underlined}} for each metric.}
    \label{tab:WikiResult1}
\end{table*}
\egroup

\subsubsection{BSWs identified by Various Strategies}
%Next, we show the list of bias sensitive words recognized using various detection strategies for both the datasets.

Table~\ref{tab:biasedWords} shows the list of BSWs obtained using different Bias Detection strategies. The words obtained from both the SOAC and SPCPD detection strategies are not obvious and differ from the manual set of words as the distribution of the terms in the training set can be arbitrary, motivating us to design detection strategies. In case of the Twitter Dataset, in Table~\ref{tab:biasedWords}, we do not show results for the `Manual List Creation' detection strategy because we do not have a manual dataset of bias sensitive words for Twitter. The simple strategy of detecting biased words using Skewed Predicted Class Probability Distribution (SPCPD) is effective and works for arbitrary classifiers as well. The words obtained from the SPCPD detection strategy are not very obvious but the prediction on sentences that contain those words have very high discrepancy than expected prediction scores. 

The BSWs obtained using the detection strategies are better than the ones obtained using the `Manual List Creation' strategy because of the high $PB_{asym}$ scores for the `Biased' classifier for different detection strategies as tabulated in Table~\ref{tab:WikiResult1}. Additionally, the $PB_{asym}$ scores for BSWs from the manual list are very low in spite of a large number of words which suggests that the classifier is not biased towards these terms. For the Twitter dataset, the $PB_{asym}$ score for the SOAC detection strategy is the lowest, but in WikiDetox where the training dataset size is large, the skewness in distribution reduces resulting in the $PB_{asym}$ scores for the SPCPD detection strategy to be the lowest.

\subsubsection{Effectiveness of various Bias Detection-Replacement Strategies}

Tables~\ref{tab:TwitterResult2},~\ref{tab:madlibsResult3} and ~\ref{tab:WikiResult1} show the hate speech detection accuracy and bias mitigation results across various bias detection and bias replacement strategies (averaged over 10 runs because of high variance) for the Twitter, Wiki Madlibs and WikiDetox datasets. `Biased' replacement strategy indicates the biased classifier, i.e., the one where no bias has been removed. For Wiki De-bias bias removal strategy, we do not have results for non-manual detection strategies since they require template creation (like `Wiki Madlibs') and dataset curation. We train the `Wiki De-bias'~\cite{dixon2017measuring} baseline using the additional data containing 3,465 non-toxic samples as part of the dataset augmentation method as proposed in the original paper. For Wiki Madlibs dataset (Table~\ref{tab:madlibsResult3}), we show results using only manual list creation but across multiple bias replacement strategies, using $pAUC$ metric only since it is meant for this combination alone. WordNet-$i$ denotes the WordNet strategy where the starting synset level for finding hypernyms was set to $i$ levels up compared to the level of the bias sensitive word. For Tables~\ref{tab:TwitterResult2} and ~\ref{tab:WikiResult1}, we use the ROC-AUC scores for comparing the general model performance while the $PB_{asym}$ metric for bias evaluation. We do not use $pAUC$ for these tables because they need manual generation of a dataset like Wiki Madlibs. Finally, note that the scores in the Table \ref{tab:TwitterResult2} and Table \ref{tab:WikiResult1} are vertically grouped based on different detection strategies which involve varying set of BSWs making the results non-comparable across the detection strategies in general.

As seen in Table~\ref{tab:madlibsResult3}, for the Wiki Madlibs dataset, Centroid Embedding strategy performs the best beating the state-of-the-art technique (Wiki De-Bias) on both ROC-AUC as well as pAUC metrics. Additionally, POS tags and WordNet-based strategies also perform better than the Wiki De-bias strategy on the $pAUC$ metric with just a minor reduction in $ROC-AUC$ accuracy. The POS tags strategy performs better because the replaced words mostly belong to the same POS tag resulting in high generalization.

For the Twitter dataset, Centroid Embedding replacement method works best across all the detection strategies and across different metrics. The overall general classification performance for all the strategies remain almost similar, though the WordNet-3 replacement strategy works best. WordNet based replacement strategy is not as effective as the $PB_{asym}$ scores gradually increase with the increase in generalization, due to the absence of slang Twitter words in WordNet. The WordNet-based replacement strategies also show decreasing $PB_{asym}$ scores as the strategy is unable to find a suitable generalized concept that is already present in the Twitter dataset because of its small vocabulary size.

The WordNet and Centroid Embedding replacement strategies work best for the WikiDetox dataset. For the WordNet strategy, the best results vary based on detection strategy, but it performs either the best or gives comparable results to the best strategy. It also has an added benefit of deciding on the level of generalization which can be tweaked as per the task. For ``manual list creation'' detection strategy, Centroid Embedding provides unexpectedly low $PB$ scores because most of the words in the list are nouns. The non-zero $PB_{mean}$ metric in case of manual-list-creation detection strategy along with POS tags replacement strategy is due to the fact that the words like `gay' are marked as Adjective instead of Noun, which is expected.

%Table~\ref{tab:madlibsResult3} shows the bias mitigation results using pAUC metric across multiple replacement strategies on the Wiki Madlibs dataset. As can be seen from the table, Centroid Embedding performs better than state-of-the-art (Wiki De-Bias) on both ROC-AUC as well as pAUC metrics. The POS tags strategy also performs better because the replaced words mostly belong to the same POS tag resulting in high generalization.

Finally, note that bias mitigation usually does not lead to any significant change in the ROC-AUC values, if performed strategically, while generally, it reduces bias in the learning.

\subsubsection{Details of WordNet-based bias removal strategy}

Table~\ref{tab:WordNetResults4} shows WordNet-based replacement examples for the manual set of words with different values of the start synset level. 

\bgroup
\begin{table}[!phtb]
    \centering
    \small
    \begin{tabular}{|l||c|c|c|c|c|c|}
        \hline
         \textbf{Original} & \multicolumn{6}{c|}{\textbf{Replaced Words}} \\
                 \cline{2-7}
         \textbf{Word} & \textbf{Level 0} & \textbf{Level 1} & \textbf{Level 2} & \textbf{Level 3} & \textbf{Level 4} & \textbf{Level 5} \\
        \hline
        \hline
        sikh & \multicolumn{2}{c|}{disciple} & follower & person & cause & entity    \\ \hline
        homosexual& homosexual    & person & cause & \multicolumn{2}{c|}{entity} & object    \\ \hline
        queer& faggot & \multicolumn{2}{c|}{homosexual} & person & cause & entity    \\ \hline
        gay& homosexual    & person & cause & \multicolumn{2}{c|}{entity} & object    \\ \hline
        straight& homosexual & person & cause & collection & entity & object    \\ \hline
        muslim& \multicolumn{3}{c|}{person} & cause & \multicolumn{2}{c|}{entity} \\ \hline
        deaf& deaf    & people & group & abstraction & entity & - \\ \hline
        latino& \multicolumn{3}{c|}{inhabitant} & person & cause & entity \\ \hline
    \end{tabular}
    \caption{WordNet based bias correction examples for the `Manual List Creation' detection strategy with different start Synset Levels}
    \label{tab:WordNetResults4}
\end{table}
\egroup

\begin{figure*}[!htb]
    \includegraphics[width=\linewidth]{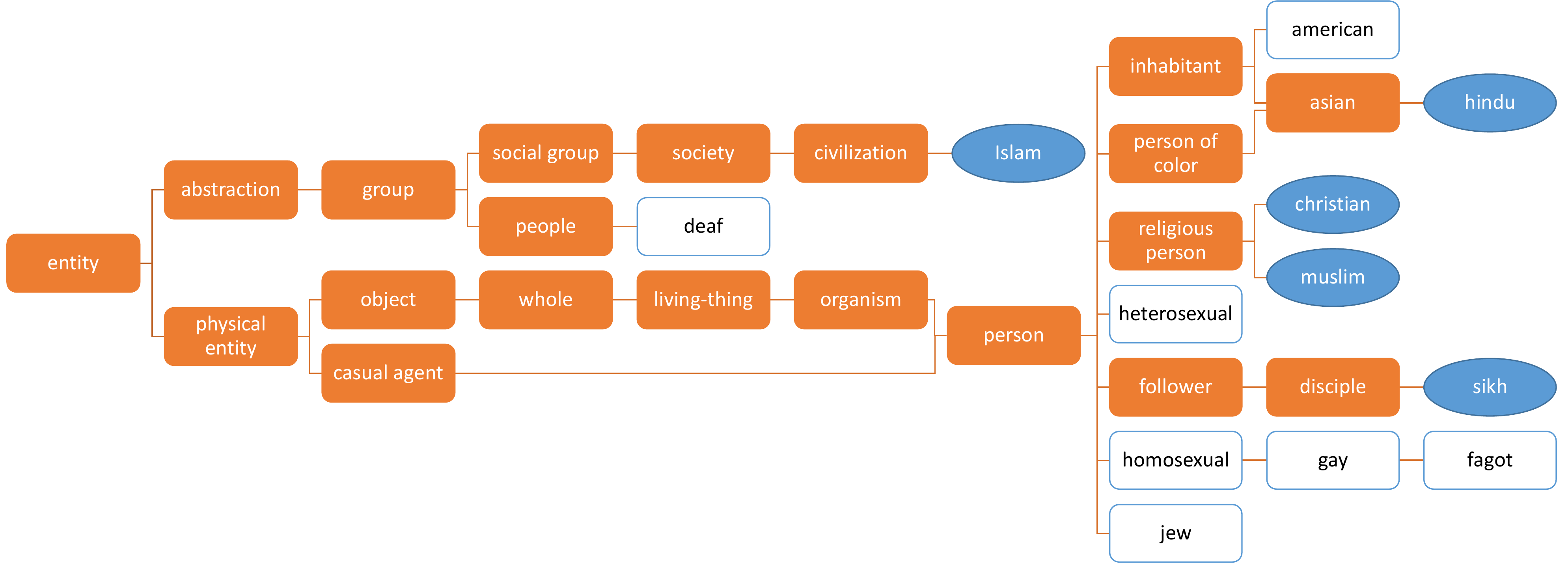}%
    \caption{Horizontal representation of the WordNet Hypernym Hierarchy for BSWs.}%
    \label{fig:wordnet-hierarcy}%
\end{figure*}

\newcommand*\boxExample{ \protect \includegraphics[scale=0.02]{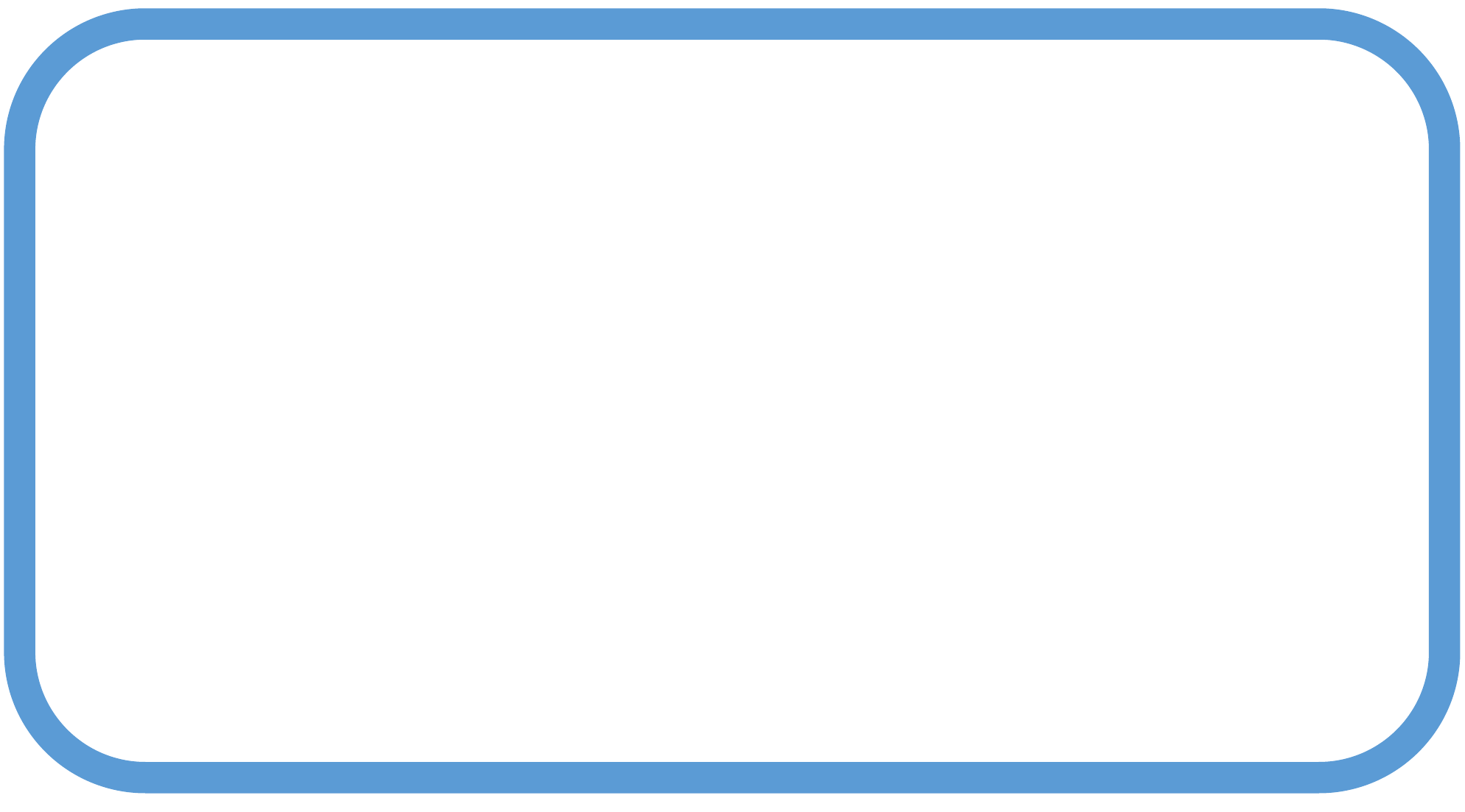}}
\newcommand*\circleExample{ \protect \includegraphics[scale=0.02]{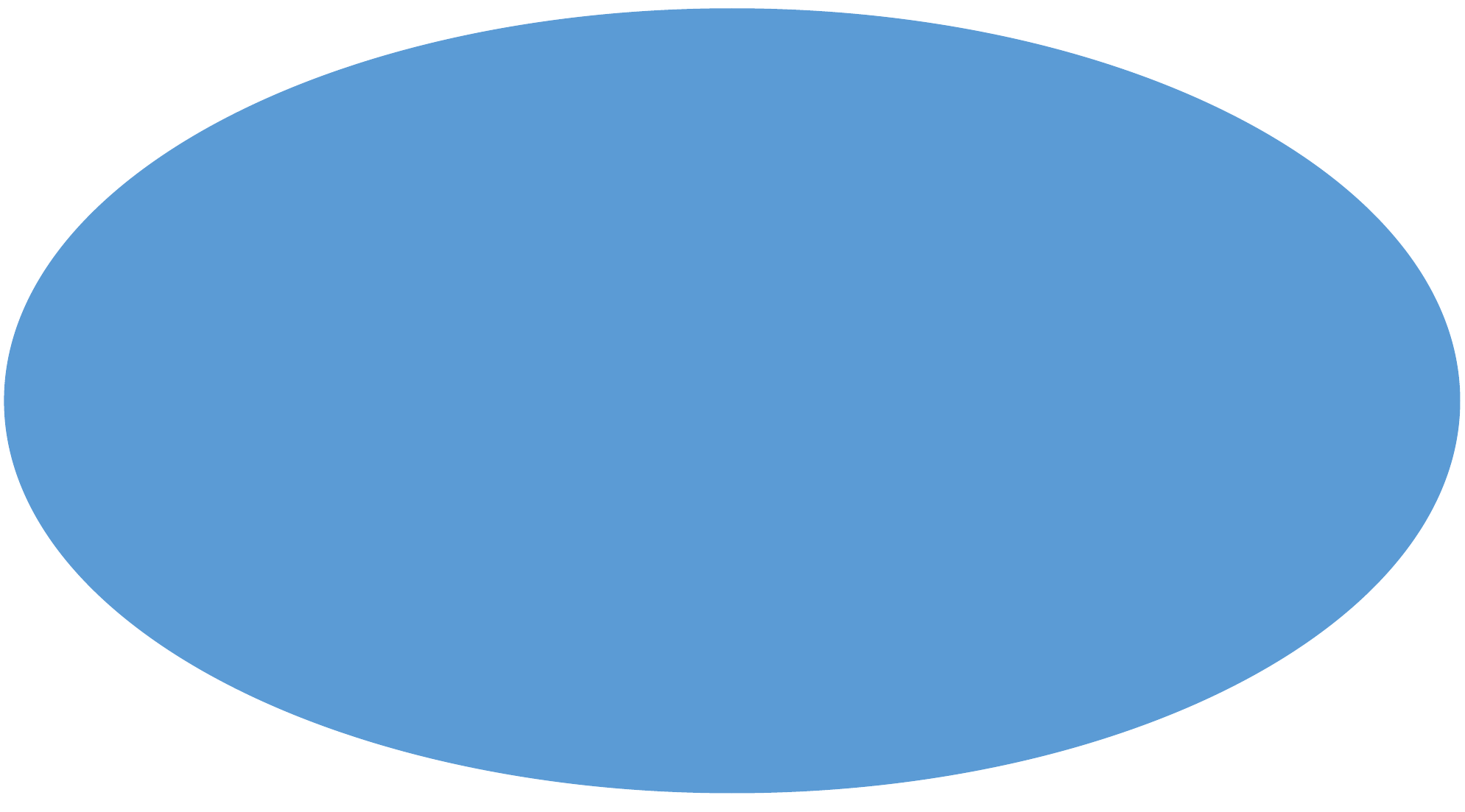}}
Fig.~\ref{fig:wordnet-hierarcy} shows details about WordNet based replacement. As we traverse from right-to-left along the edges, we obtain generalizations for each of the words. The figure highlights two sets of similar words where one set (denoted by~\boxExample~) is highly prone towards getting stereotypically biased, while all the highlighted terms (denoted by~\circleExample~) represent ``a group of people with specific attributes''. Each node represents the term which can be used for BSW replacement. In the case of multiple generalizations, the selection was made randomly.

Wordnet-based strategy often leads to the best results across our datasets. However, the results could be improved further using an appropriate normalization scale that captures generalization to different levels. For example, \textit{Hindu} requires three levels of generalization, while \textit{Muslim} requires two (see Fig.~\ref{fig:wordnet-hierarcy}). This requires a generalization scheme that considers the inconsistency in the sibling nodes and performs differential normalization.

%\textit{Wordnet-based Knowledge generalization does not perform the best. Centroid Embedding performs the best. Why?} Wordnet-based strategies have results comparable to the best strategy a lot of times. One of the reasons that we identified why it did not perform the best was due to the need for a differential normalization scale in each of the branches. For example, \textit{Hindu} requires 3 levels of generalization, while \textit{Muslim} requires 2 (see Fig. \ref{fig:wordnet-hierarcy}). This requires a generalization scheme that considers the inconsistency in the sibling nodes and performs differential normalization. Additionally, Wordnet-based strategies have scores close to the best performing technique most of the times suggesting an ideal fit for a practical use case. Centroid Embedding, on the other hand, works better because most the similar words that lie in different levels in the Wordnet hierarchy, they usually are amongst the nearest K neighbors.

\subsubsection{Case Studies}

In order to evaluate the effectiveness of the Bias Detection strategy in identifying BSWs, we perform an experiment on a publicly accessible classifier, the Perspective API~\footnote{https://www.perspectiveapi.com/}. We leverage the vocabulary of size 400k of GloVe vectors and obtain hate (toxicity) scores for each word in the vocabulary and rank them using the Skewed Predicted Class Probability Distribution Bias Detection strategy. This allows us to identify the set of bias sensitive words allowing us to create examples to test the quality of the obtained words. Some of these examples are tabulated in Table~\ref{tab:examples} where the API gives incorrect scores. The obtained examples show the effectiveness of the obtained BSWs.

Further, Table~\ref{tab:debiasedExamples} shows some examples which were wrongly predicted as being ``hateful'' by the original `Biased' classifier and the corresponding prediction scores after bias removal using the above strategies. These simple manually created examples clearly show the efficacy of our proposed method.

\begin{table}[!htbp]
    \centering
    \small
    \begin{tabular}{|p{5cm}|p{1.0cm}|p{1.0cm}|}
        \hline
        \multirow{3}{*}{\textbf{Examples}} & \multicolumn{2}{c|}{\textbf{Toxicity}} \\
         & \multicolumn{2}{c|}{\textbf{Predictions}} \\
        \cline{2-3}
        &  \textbf{before bias removal} & \textbf{after bias removal} \\ \hline
        are you a \textbf{nerd} ?  & 0.928 & 0.294 \\ \hline
        he hit the \textbf{donkey} & 0.289 & 0.021 \\ \hline
        he is a \textbf{gay}  & 0.786 & 0.493 \\ \hline
        One of John brothers was \textbf{homosexual} while the other is a \textbf{black transgender} & 0.690 & 0.048 \\ \hline
        katrina is a \textbf{female african} who is \textbf{sikh} & 0.782 & 0.535 \\ \hline
    \end{tabular}
    \caption{Examples for predictions on sentences before and after the classifier had undergone the bias correction stage using the Centroid Embedding strategy for the manually constructed sentences containing identified BSWs. Bias Sensitive Words are marked in bold. The ideal prediction probabilities in all the cases should be less than 0.5}
    \label{tab:debiasedExamples}
\end{table}

%In spite of improvements across different metrics, we try to perform qualitative analysis by manually constructing examples containing BSWs in order to verify the reduction in bias.  We also show results in 

%\subsection{Inference}

\subsubsection{Unigrams versus higher order ngrams}
In this paper, we experimented with unigrams only. From our experiments, we observed that the SB bias in unigrams propagates to word n-grams and char-n-grams as well. E.g., SB towards a word `women' propagates to the char n-grams `wom, omen, wome' as the distribution of char n-grams is similar to that of the corresponding word and also similar to the word n-grams, \textit{women are}, \textit{womens suck}, etc. Thus, considering unigrams is reasonably sufficient. %We also observe that results are similar with the use of LSTM (that give more importance to word-sequence than CNN) based model architecture. 
That said, the PB metrics rely on the probability scores and thus can be easily used for bias sensitive \emph{n-grams} as well.
\section{Conclusions}
\label{sec:conclusions}

In this paper, we discussed the problem of bias mitigation for the hate speech detection task. We proposed a two-stage framework to address the problem. We proposed various heuristics to detect bias sensitive words. We also proposed multiple novel knowledge generalization based strategies for bias sensitive word replacement which can be extended to other short-text classification tasks as well. Using two real-world datasets, we showed the efficacy of the proposed methods. We also empirically show that the data-correction based removal techniques can reduce the bias without reducing the overall model performance.

We also demonstrate the effectiveness of these strategies by performing the experiment on Google Perspective API and perform a qualitative analysis to support our observations using multiple examples.
\bibliographystyle{ACM-Reference-Format}
\balance 
\bibliography{references}

\end{document}